\documentclass{bmvc2k}
\usepackage{amsfonts}
\usepackage{algorithm}
\usepackage{algpseudocode}
\usepackage{booktabs}
\usepackage{multirow}
\usepackage{graphicx}
\usepackage{mathrsfs}
\usepackage{verbatim}

\newcommand{\argmin}{\mathop{\rm arg~min}\limits}

\title{LagrangeGS: Non-Conservative Lagrangian System on Dynamic 3D Gaussian Splatting}

\addauthor{Shogo Sato}{shg.sato@ntt.com}{1}
\addauthor{Takuhiro Kaneko}{takuhiro.kaneko@ntt.com}{1}
\addauthor{Shoichiro Takeda}{shoichiro.takeda@ntt.com}{1}
\addauthor{Tomoyasu Shimada}{tomoyasu.shimada@ntt.com}{1}
\addauthor{Riku Inoue}{riku.inoue@ntt.com}{1}
\addauthor{Kazuhiko Murasaki}{kazuhiko.murasaki@ntt.com}{1}
\addauthor{Ryuichi Tanida}{ryuichi.tanida@ntt.com}{1}

\addinstitution{
Human Informatics Laboratories\\
NTT, Inc.\\
Kanagawa, Japan
}

\runninghead{S.Sato et al.}{LagrangeGS}

\begin{document}

\maketitle

\begin{figure}[ht]
\centering
\includegraphics[width=1.\linewidth]{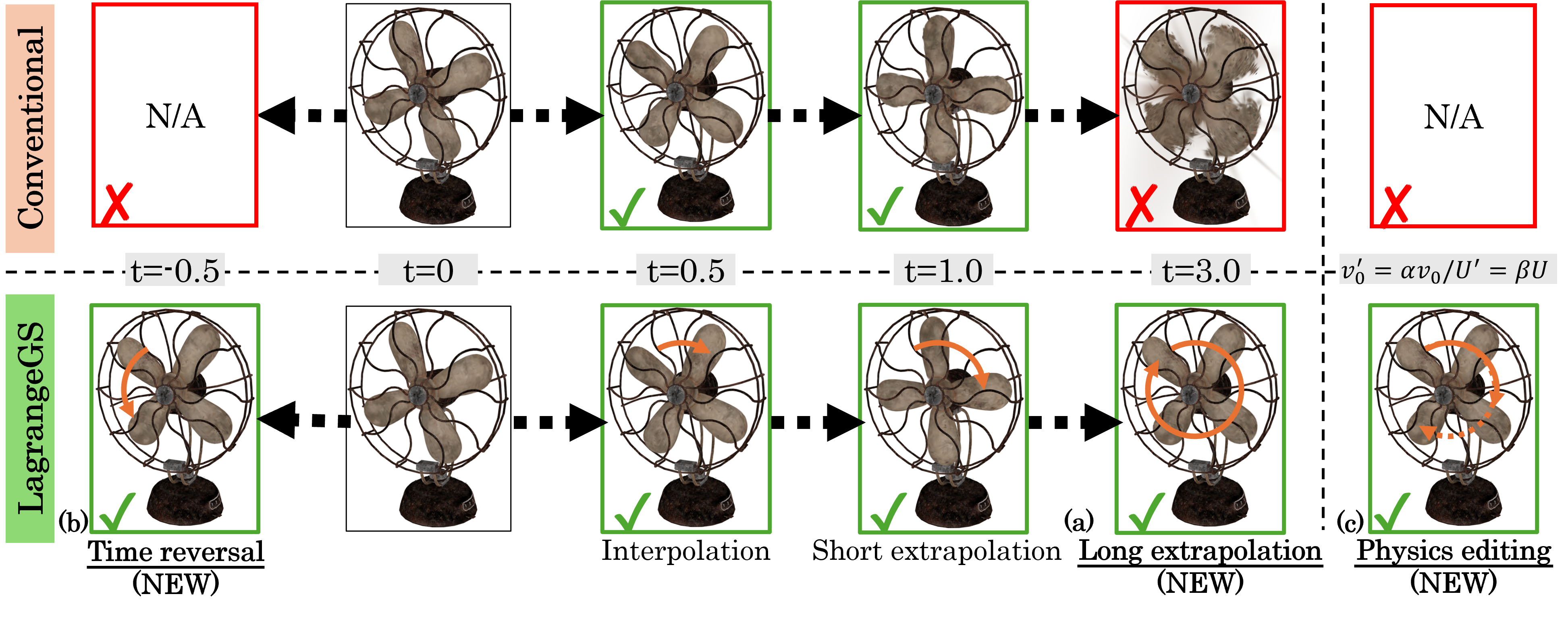}
\vspace{-4mm}
\caption{\label{fig1} LagrangeGS formulates dynamic scenes as a non-conservative Lagrangian system parameterized by potential energy $U$, initial velocity $v_0$, and non-conservative forces $Q$, enabling (a) long-term extrapolation, (b) time reversal, and (c) physics-based editing.}
\vspace{-5mm}
\end{figure}

\begin{abstract}
Dynamic 3D Gaussian Splatting (3DGS) achieves photorealistic reconstruction of time-varying scenes, and recent physics-aware extensions improve extrapolation by explicitly predicting velocity fields. However, these extensions merely fit vector fields to visual deformations without satisfying Lagrangian mechanics, leading to three major issues: (i) physically inconsistent trajectories, (ii) lack of time-reversibility, and (iii) geometric collapse during long-term extrapolation. In this paper, we propose LagrangeGS, which formulates dynamic 3DGS as a non-conservative Lagrangian system. While this Lagrangian formulation fundamentally solves (i), a direct application of general LNNs to dynamic 3DGS requires a large velocity-Hessian inversion for millions of Gaussian particles. To overcome this computational bottleneck, we approximate the velocity-Hessian as an identity matrix, decoupling particle dynamics for computational tractability. For (ii), we restrict the non-conservative forces to be explicitly time independent, enabling consistent backward integration. Finally, to address (iii), we introduce local rigid alignment that regularizes particle trajectories. Extensive evaluations on dynamic scene benchmarks demonstrate that LagrangeGS enables stable long-term extrapolation, consistent time reversal, and counterfactual physics-based editing without retraining.
\end{abstract}

\section{\label{sec1}Introduction}
Physical modeling of dynamic 3D scenes is a key enabler for downstream applications such as robot motion planning and embodied artificial intelligence (AI). These advanced applications demand not only visual fidelity but also physics manipulations: recovering past trajectories from current states (time reversal) and simulating "what-if" scenarios under altered environments (counterfactual physics-based editing). The emergence of Neural Radiance Fields (NeRF)~\cite{mildenhall2021nerf} and 3D Gaussian Splatting (3DGS)~\cite{kerbl20233d} has driven rapid progress in static 3D scene reconstruction. While their dynamic extensions successfully render time-varying scenes within the training window (interpolation)~\cite{pumarola2021d, li2023dynibar, wu20244d, yang2024deformable}, they operate in a purely visual domain without physics priors. Consequently, performance beyond the training window (extrapolation) is limited, and physics manipulations remain ill-defined. To bridge this gap, recent physics-aware extensions explicitly predict velocity~\cite{li2025freegave} or acceleration~\cite{li2025trace} fields. However, these extensions merely fit vector fields to visual deformations without satisfying Lagrangian mechanics. This fundamental limitation leads to three major issues: (i) physically inconsistent trajectories, (ii) lack of time-reversibility, and (iii) geometric collapse during long-term extrapolation.

To address these issues, we propose LagrangeGS, a Lagrangian Neural Network (LNN)~\cite{cranmer2020lagrangian} based framework that formulates dynamic 3DGS as a non-conservative Lagrangian system explicitly parameterized by potential energy, initial velocity, and non-conservative forces. This Lagrangian formulation fundamentally solves (i) and formally defines physics manipulations (Figure~\ref{fig1}). However, the direct application of conventional LNN~\cite{xiao2024generalized} to dynamic 3DGS requires an intractable velocity-Hessian inversion for calculating the velocity of millions of Gaussian particles during forward integration. To overcome this computational bottleneck, we approximate the velocity-Hessian as an identity matrix to avoid computing its inverse. This operation physically corresponds to assigning a uniform unit mass to all Gaussian particles and decoupling their inertial dynamics. To address (ii), we restrict the non-conservative forces to be explicitly time independent, enabling consistent time reversal. We refer to this architecture, which combines the decoupled particle dynamics and the time-independent system, as the Time-independent Decoupled LNN (TD-LNN). Finally, to address (iii), we propose a local rigid alignment (LRA), which binds the decoupled particles into locally rigid structures via a Kabsch solver~\cite{kabsch1976solution, umeyama1991least}.

We validate LagrangeGS on the Dynamic Object~\cite{li2023nvfi} and Dynamic Indoor Scene~\cite{li2023nvfi} benchmarks across short-term extrapolation, long-term extrapolation, time reversal, and counterfactual physics-based editing. Our contributions are summarized as follows:
\begin{itemize}
\item We propose LagrangeGS, which formulates dynamic 3DGS as a non-conservative Lagrangian system. This Lagrangian formulation fundamentally solves physically inconsistent trajectories and defines physics manipulations.
\item We introduce two key architectural designs: (1) TD-LNN, which approximates the velocity-Hessian as an identity matrix for computational tractability and employs a time-independent system for consistent time reversal; and (2) LRA, which binds the particles into locally rigid structures to prevent geometric collapse.
\item We demonstrate that LagrangeGS enables stable long-term extrapolation, consistent time reversal, and counterfactual physics-based editing without retraining.
\end{itemize}

\section{\label{sec2}Related Work}
\textbf{Dynamic 3D Scene Representation} Various models have been proposed for dynamic 3D scene representation, ranging from NeRF-based models~\cite{pumarola2021d, li2023dynibar} to 3DGS-based models~\cite{wu20244d}. DefGS~\cite{yang2024deformable} introduces Gaussian deformation networks to handle scene dynamics, yet they operate in a purely visual domain without physics priors, leading to limited extrapolation performance. To enable extrapolation, recent approaches introduce physics priors into dynamic 3DGS through two main strategies: vector fields and external simulation. Models adopting the first strategy predict velocity fields in NVFi~\cite{li2023nvfi} and FreeGave~\cite{li2025freegave}, or acceleration fields in TRACE~\cite{li2025trace}. Other methods apply continuous-time regularization via Neural ODEs~\cite{chen2018neural} (ParticleGS~\cite{quan2025particlegs}) or Hamiltonian formulations~\cite{greydanus2019hamiltonian} (NeHaD~\cite{qin2025neural}). However, these models impose physical constraints by merely fitting vector fields to visual deformations. Since they do not satisfy Lagrangian mechanics, they suffer from physically inconsistent trajectories and geometric collapse during long-term extrapolation as well as provide no theoretical foundation for physics manipulations. The second strategy integrates the Material Point Method (MPM)~\cite{sulsky1994particle, hu2019chainqueen} into the optimization loop~\cite{li2023pac}, as seen in PhysGaussian~\cite{xie2024physgaussian} and Physics3D~\cite{liu2024physics3d}. While these simulation-based methods satisfy Lagrangian mechanics, embedding an MPM solver within the optimization loop is computationally expensive, and often leads to vanishing or exploding gradients. In contrast, LagrangeGS requires no external simulator, yet satisfies Lagrangian mechanics.\smallskip\\ 
\textbf{Neural Dynamics Representations} Modeling continuous state evolution via neural differential equations, such as Neural ODEs~\cite{chen2018neural}, has gained significant traction. HNNs~\cite{greydanus2019hamiltonian} embed symplectic structures to conserve total energy, yet they require canonical coordinates that are often difficult to rigorously define for complex 3D scenes. LNNs~\cite{cranmer2020lagrangian} relax this requirement by learning the Lagrangian $\mathcal{L}(q, \dot{q})$ directly in generalized coordinates, though vanilla LNNs are strictly limited to conservative systems. To accommodate non-conservative behaviors, models like Deep Lagrangian Networks (DeLaN)~\cite{lutter2019deep} and the Generalised LNN (G-LNN)~\cite{xiao2024generalized} introduce explicit external forces to Lagrangian mechanics. Crucially, G-LNN's standard formulation introduces two fatal flaws for 3DGS applications: first, it necessitates the inversion of a velocity-Hessian matrix $\partial^2\mathcal{L}/\partial\dot{q}^2$ at each integration step, making it computationally intractable for the millions of Gaussian particles of a standard dynamic 3DGS scene. Second, its explicit time-dependent force $F(q,\dot{q},t)$, making consistent time reversal mathematically ill-defined. In our model, we approximate the velocity-Hessian as an identity matrix to avoid the intractable computation. Additionally, we restrict the non-conservative forces to be explicitly time independent $Q(q,\dot{q})$, enabling exact backward integration.

\section{\label{sec3} Methods}
\subsection{\label{sec3-1} Preliminary}
\textbf{Task formulation.} The goal of dynamic 3D scene rendering is to synthesize photorealistic images of a time-varying scene from arbitrary novel viewpoints. The input consists of a set of RGB video frames $\{I^1_\tau, I^2_\tau, \ldots, I^N_\tau\}$ captured by $N$ synchronized cameras at discrete time steps such as $\tau = 0, 0.01, \ldots, \tau_{\rm{max}} \in \mathbb{R}_{+}$, where $\tau_{\rm{max}}$ denotes the maximum time step of the observed sequence. The system aims to render images at an arbitrary target time $t' \in \mathbb{R}$ and novel viewpoint $v'$. Conventional benchmarks primarily evaluate temporal interpolation ($t' \in [0, \tau_{\rm{max}}]$) and short-term extrapolation ($t' \in (\tau_{\rm{max}}, \tau_{\rm{max}}+\Delta]$ with a small $\Delta$). In this work, we additionally formulate three challenging physics manipulations: (i) long-term extrapolation ($t' \in (\tau_{\rm{max}}, 3\tau_{\rm{max}}]$), (ii) time reversal ($t' \in [-\tau_{\rm{max}}, 0)$), and (iii) counterfactual physics-based editing, where the underlying dynamics are explicitly modified by scaling physical parameters such as potential energy, initial velocity, and non-conservative forces.\smallskip\\
\textbf{3D Gaussian Splatting~\cite{kerbl20233d}.} Following vanilla 3DGS, a static scene at the initial time $\tau = 0$ is represented as a set of 3D Gaussian particles, $\mathcal{G}_{0} = \{g^1_0, g^2_0, \dots, g^M_0\}$. Each Gaussian $g^j_0$ is parameterized by its mean position $q^j_0 \in \mathbb{R}^{3}$, a covariance matrix derived from a quaternion $r^j_0 \in \mathbb{R}^{4}$, a scaling factor $s^j_0 \in \mathbb{R}^{3}$, an opacity $\sigma^j_0 \in \mathbb{R}$, and a color coefficient $c^j_0 \in \mathbb{R}^{d_{c}}$, where $d_c$ is the dimension of the color space. The Gaussian set is optimized to reconstruct the initial frames $\{I^1_0, I^2_0, \ldots, I^N_0\}$ by minimizing the photometric loss with the Structural Similarity Index Measure (SSIM):
\begin{equation}
\label{eq1++}
\mathcal{L}_\text{img} = \sum_{i=1}^{N} (1-\lambda)\lVert \hat{I}^i_0 - I^i_0\rVert_1 + \lambda\,(1 - \mathrm{SSIM}(\hat{I}^i_0, I^i_0)),
\end{equation}
where $\hat{I}^i_0$ and $I^i_0$ are the rendered image from the $i$-th camera at the initial frame and its corresponding ground truth. \smallskip\\
\textbf{Deformation GS (DefGS)~\cite{yang2024deformable}.} To model scene dynamics, a deformation network $f_d$ is learned to predict deviations from the initial state ($\tau=0$). Specifically, given a time $t'$, the network predicts the position displacement $\delta q^j_{t'}$, quaternion displacement $\delta r^j_{t'}$, and scaling factor displacement $\delta s^j_{t'}$ for each Gaussian particle, yielding the deformed state directly
\begin{equation}
\label{eq1}
q^j_{t'} = q^j_0 + \delta q^j_{t'}, \quad r^j_{t'} = r^j_0 + \delta r^j_{t'}, \quad s^j_{t'} = s^j_0 + \delta s^j_{t'},
\end{equation}
while opacity $\sigma^j_{t'}$ and color $c^j_{t'}$ are typically kept time-invariant ($\sigma^{j}_{t'} = \sigma^{j}_0,\ c^{j}_{t'} = c^{j}_0$). The deformation network $f_d$ is also trained with the photometric loss for all time frames. Although DefGS reconstructs the observed motion faithfully, it relies solely on fitting visual deformations without physics priors, leading to limited extrapolation performance and ill-defined physics manipulations.\smallskip\\
\textbf{Generalized Lagrangian Neural Networks (G-LNN)~\cite{xiao2024generalized}.} To satisfy Lagrangian mechanics, we apply LNN to dynamic 3DGS. Given $M$ particles with generalized coordinates $q = [q^1_t, q^2_t, \dots, q^M_t]^T \in \mathbb{R}^{3M}$ and generalized velocities $\dot{q} = [\dot{q}^1_t, \dot{q}^2_t, \dots, \dot{q}^M_t]^T \in \mathbb{R}^{3M}$ at an arbitrary continuous time $t \in \mathbb{R}$, a standard LNN learns a Lagrangian $\mathcal{L}(q, \dot{q})$: 
\begin{equation}
\label{eq1+}
\mathcal{L}(q, \dot{q}) = T(q, \dot{q}) - U(q),
\end{equation}
where $T$ and $U$ denote the kinetic and potential energies, respectively. The system state evolves through the Euler-Lagrange equation:
\begin{equation}
\label{eq2}
\frac{d}{dt} \left( \frac{\partial \mathcal{L}}{\partial \dot{q}} \right) - \frac{\partial \mathcal{L}}{\partial q} = 0,
\end{equation}
which enforces conservative dynamics. The G-LNN introduces a learned non-conservative force $F(q, \dot{q}, t)$ to the right-hand side to accommodate dissipation and external interactions:
\begin{equation}
\label{eq3}
\frac{d}{dt} \left( \frac{\partial \mathcal{L}}{\partial \dot{q}} \right) - \frac{\partial \mathcal{L}}{\partial q} = F(q, \dot{q}, t).
\end{equation}
Expanding the total time derivative using the chain rule yields $\frac{\partial^2 \mathcal{L}}{\partial \dot{q}^2} \ddot{q} + \frac{\partial^2 \mathcal{L}}{\partial \dot{q} \partial q} \dot{q} - \frac{\partial \mathcal{L}}{\partial q} = F(q, \dot{q}, t)$. Solving for the physical acceleration $\ddot{q} \in \mathbb{R}^{3M}$ requires isolating it:
\begin{equation}
\label{eq3+}
\ddot{q} = \left( \frac{\partial^2 \mathcal{L}}{\partial \dot{q}^2} \right)^{-1} \left[ \frac{\partial \mathcal{L}}{\partial q} - \frac{\partial^2 \mathcal{L}}{\partial \dot{q} \partial q} \dot{q} + F(q, \dot{q}, t) \right].
\end{equation}
For a general LNN, the kinetic term $T(\mathbf{q},\dot{\mathbf{q}})$ may couple the velocities of different degrees of freedom. Consequently, the velocity-Hessian $\partial^2 L/\partial\dot{\mathbf{q}}^2$ is not necessarily diagonal even when Cartesian coordinates are used. For $M$ Gaussian particles, directly evaluating Eq.~(6) may therefore require solving a coupled $3M$-dimensional system at every integration step, which is impractical for dynamic 3DGS containing millions of particles.

\begin{figure}
\centering
\includegraphics[width=1.0\linewidth]{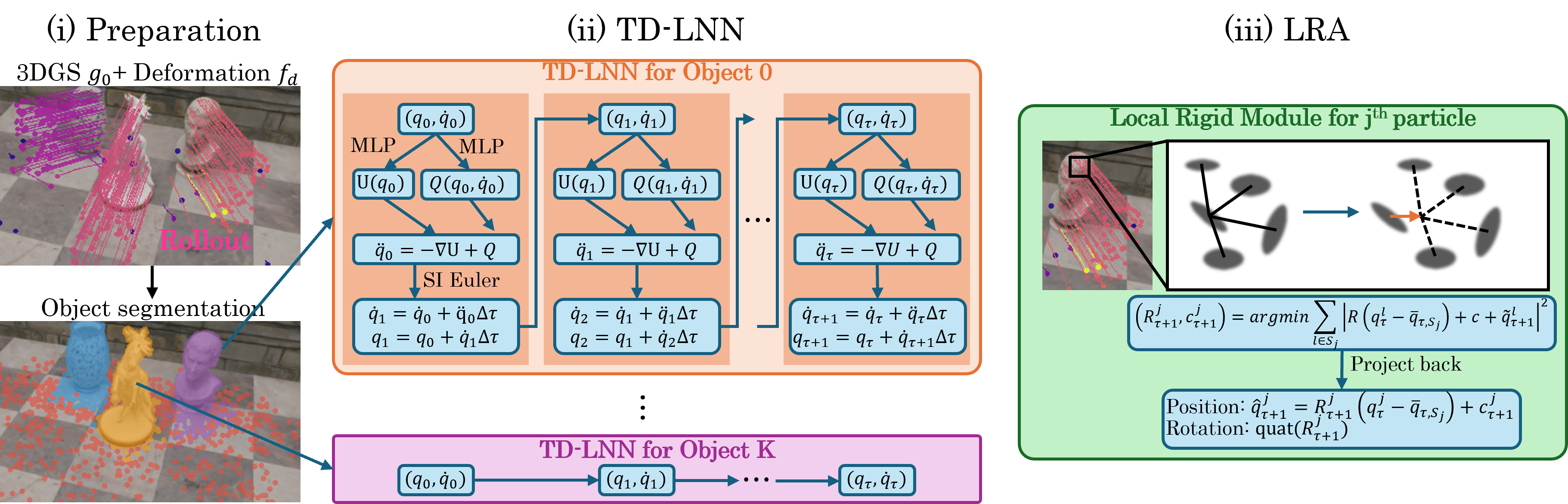}
\vspace{-3mm}
\caption{\label{fig2} Overview of LagrangeGS. (i) Gaussian particles are partitioned into distinct objects via pre-trained model trajectory. (ii) TD-LNN computes per-particle accelerations $\ddot{q} = -\nabla_q U + Q$, combining the decoupled particle dynamics for avoiding intractable computation
and the time-independent system for exact backward integration. (iii) To prevent geometric collapse, LRA binds the decoupled particles into locally rigid structures.}
\vspace{-5mm}
\end{figure}

\subsection{\label{sec3-2} LagrangeGS}
\textbf{Overall architecture.} 
The overall architecture of LagrangeGS is illustrated in Figure~\ref{fig2}. Starting from an initial 3D Gaussian representation $\mathcal{G}_0$ at $\tau=0$, we (i) extract the per-particle trajectories $\{q^{j}_\tau\}$ from the pre-trained $f_d$ and partition Gaussian particles into objects based on the trajectories, (ii) distill the per-object trajectories into an TD-LNN with  potential energy, initial velocity, and non-conservative forces, and (iii) align the candidate positions predicted by the TD-LNN with LRA. Finally, we refine both the TD-LNN parameters and the Gaussian parameters with the photometric loss. At inference time, per-Gaussian rotations are derived from the Kabsch rotations, while scaling, opacity, and color remain time-invariant.\smallskip\\
\textbf{Time-independent Decoupled LNN (TD-LNN).} To avoid an intractable velocity-Hessian inversion for millions of Gaussian particles, we assign a uniform unit mass to all Gaussian particles and decouple their inertial dynamics to simplify the kinetic energy as $T(\dot{q}) = \frac{1}{2} ||\dot{q}||^{2}$. Substituting this prior into the Lagrangian definition (Eq.~\ref{eq1+}), the momentum term simplifies to $\frac{\partial \mathcal{L}}{\partial \dot{q}} = \dot{q}$. Consequently, the velocity-Hessian $\frac{\partial^{2}\mathcal{L}}{\partial \dot{q}^{2}}$ rigorously reduces to the identity matrix $I$, and the cross-Hessian $\frac{\partial^{2}\mathcal{L}}{\partial \dot{q} \partial q} = 0$. Furthermore, the potential term derivative becomes $\frac{\partial \mathcal{L}}{\partial q} = -\nabla_{q} U(q)$. Unlike standard G-LNNs, we restrict our non-conservative force to be explicitly time independent, such that $F(q, \dot{q}, t) = Q(q, \dot{q})$. By substituting these reduced terms back into Eq.~\ref{eq3+}, the acceleration for each particle $j$ transparently reduces to a closed form:
\begin{equation}
\label{eq4}
\ddot{q} = -\nabla_{q} U(q) + Q(q, \dot{q}).
\end{equation}
This formulation depends only on a scalar potential $U:\mathbb{R}^{3M}\to\mathbb{R}$ and a time-independent non-conservative force $Q:\mathbb{R}^{3M}\times\mathbb{R}^{3M}\to\mathbb{R}^{3M}$. Crucially, it entirely bypasses the intractable full-system matrix inversion. Both $U$ and $Q$ are parameterized by MLPs~\cite{cranmer2020lagrangian}; the last layer of $Q$ is zero-initialized so training starts from a purely conservative baseline ($Q \equiv 0$) and learns the forcing correction incrementally. A single $U$ is shared across all particles within an object. This structural choice allows the TD-LNN to easily scale to millions of Gaussian particles. Each particle's state is time-integrated independently with the semi-implicit Euler scheme:
\begin{equation}
\label{eq5}
\dot{q}^j_{\tau+1} = \dot{q}^j_\tau + \ddot{q}^j_\tau \Delta \tau, \qquad q^j_{\tau+1} = q^j_\tau + \dot{q}^j_{\tau+1} \Delta \tau.
\end{equation}
Because $Q(q, \dot{q})$ has no explicit time dependence, exact backward integration $(q^j_0, \dot{q}^j_0) \mapsto (q^j_{-\tau_{\rm{max}}}, \dot{q}^j_{-\tau_{\rm{max}}})$ becomes mathematically well-posed. \smallskip\\
\textbf{Local Rigid Alignment (LRA).} To prevent the geometric collapse during long-term extrapolation and compensate for the decoupled dynamics, we introduce a LRA that binds the decoupled particles into locally rigid structure. First, we define a local rigid support $\mathcal{S}_j$ for each Gaussian particle $j$ using its $k$-nearest neighbors in an initial state ($\tau=0$). At rollout time from $\tau$ to $\tau+1$, we obtain the intermediate candidate positions $\{\tilde{q}^{j}_{\tau+1}\}$ from the TD-LNN. For each particle $j$, we structurally align these candidates onto a locally rigid motion of its support at time $\tau$ via a differentiable Kabsch fit:
\begin{equation}
\label{eq6}
(R^{j}_{\tau+1}, c^{j}_{\tau+1}) = \argmin_{R \in SO(3), c \in \mathbb{R}^3} \sum_{l \in \mathcal{S}_j} \bigl\lVert R\, (q^{l}_{\tau} - \bar{q}_{\tau, \mathcal{S}j}) + c - \tilde{q}^{l}_{\tau+1} \bigr\rVert^{2},
\end{equation}
where $\bar{q}_{\tau, \mathcal{S}_j}$ is the centroid of the support at time $\tau$. The finalized physical position for particle $j$ at time $\tau+1$ is explicitly given by $q^{j}_{\tau+1} = R^{j}_{\tau+1}\, (q^{j}_{\tau} - \bar{q}_{\tau, \mathcal{S}_j}) + c^{j}_{\tau+1}$. To maintain visual consistency, its rotation is updated by extracting the unit quaternion from $R^{j}_{\tau+1}$ and applying it to the rotation $r^{j}_{\tau+1}$. Because the Singular Value Decomposition (SVD)-based Kabsch solver is fully differentiable, gradients from the photometric loss can propagate to the TD-LNN-integrated candidate positions of the entire support. This local alignment prevents geometric collapse while smoothly accommodating global continuous deformations.\smallskip\\
\textbf{Optimization.} LagrangeGS is trained in three sequential stages.\\
(Stage 1) Trajectory distillation. We minimize the rollout Mean Squared Error (MSE) between the integrated trajectories $\{q^{j}_\tau\}$ predicted by the TD-LNN and the pre-trained source trajectories $\{q'^{j}_\tau\}$ extracted from $f_d$:
\begin{equation}
\label{eq7}
\mathcal{L}_\text{distil} = \frac{1}{M\tau_{\rm{max}}} \sum_{\tau=1}^{\tau_{\rm{max}}} \sum_{j=1}^{M} \bigl\lVert q^{j}_\tau - {q'}^{j}_\tau \bigr\rVert^{2} + \lambda_Q \mathbb{E}\left[\lVert Q \rVert^{2}\right],
\end{equation}
The second term is a small $\ell_2$ penalty that forces the dynamics onto the conservative potential $U$ wherever possible, ensuring the non-conservative force $Q$ only activates for genuine dissipation or external interactions. The initial velocity $\dot{q}^{j}_0$ is a learnable parameter initialized by the source trajectories.\\
(Stage 2) TD-LNN fine-tuning. We freeze all components except $(U, Q, \{\dot{q}^{j}_0\})$ and minimize the photometric loss (Eq.~\ref{eq1++}). Crucially, this stage allows LagrangeGS to systematically escape the overfitting of the source trajectories. Because the Kabsch alignment (Eq.~\ref{eq6}) is differentiable, gradients flow from the photometric loss directly back to the TD-LNN parameters, supporting the model to discover true physical dynamics rather than merely memorizing pre-trained deformations. Truncated Backpropagation Through Time (TBPTT) limits gradient propagation over long rollout sequences, while spatial particle subsampling randomly samples Gaussian particles during training to reduce memory consumption.\\
(Stage 3) Gaussian parameter finetuning. Finally, we freeze the TD-LNN parameters $(U, Q, \{\dot{q}^{j}_0\})$ and refine only the Gaussian parameters $(q^{j}_0, r^{j}_0, s^{j}_0, \sigma^{j}_0, c^{j}_0)$ using the photometric loss. This strategically absorbs residual visual errors without corrupting the rigorously learned physics.

\begin{figure}
\centering
\includegraphics[width=1.0\linewidth]{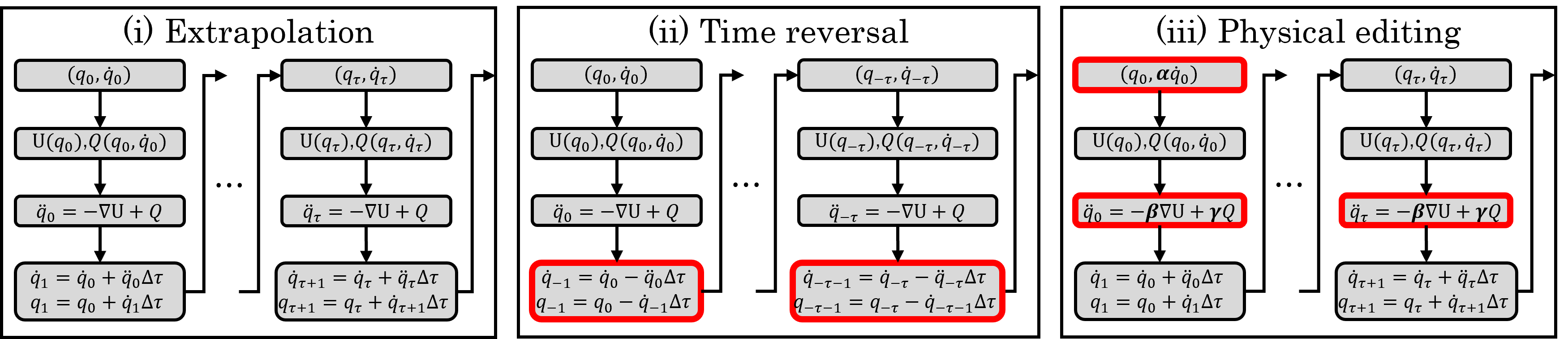}
\vspace{-3mm}
\caption{\label{fig3} Overview of the physics manipulations. (i) Extrapolation: future states ($0<\tau$) are computed by forward integration. (ii) Time reversal: Past states ($\tau < 0$) are recovered by backward integration using a negative time step $-\Delta \tau$. (iii) Counterfactual physics-based editing: Dynamic properties are manipulated by scaling the initial velocity ($\alpha$), learned potential ($\beta$), and time-independent non-conservative force ($\gamma$).}
\vspace{-5mm}
\end{figure}

\subsection{\label{sec3-3} Physics manipulations}
The Lagrangian formulation in LagrangeGS enables long-term extrapolation and the physics manipulations at inference time without retraining:\smallskip\\
\textbf{Long-term extrapolation.} For time steps beyond the training window ($t' > \tau_{\rm{max}}$), we directly extend the forward integration of Eq.~\ref{eq5} with the potential $U(q)$ and non-conservative force $Q(q, \dot{q})$. Because $Q$ has no explicit time dependence, it preserves time-translation symmetry and supporting stable long-term extrapolation.\smallskip\\
\textbf{Time reversal.} To reconstruct past states of the scene ($t' < 0$), we solve the system backward in time by reversing the semi-implicit Euler integration with a negative time step $-\Delta \tau$, as shown in Figure~\ref{fig3}:
\begin{equation}
\label{eq8}
\dot{q}^j_{\tau-1} = \dot{q}^j_\tau - \ddot{q}^j_\tau \Delta \tau, \qquad q^j_{\tau-1} = q^j_\tau - \dot{q}^j_{\tau-1} \Delta \tau.
\end{equation}
Because the system is time independent, backward integration is mathematically well-posed, enabling exact state recovery $(q^j_0, \dot{q}^j_0) \mapsto (q^j_{-\tau_{\rm{max}}}, \dot{q}^j_{-\tau_{\rm{max}}})$.\smallskip\\
\textbf{Counterfactual physics-based editing.} As shown in Figure~\ref{fig3}, we can directly manipulate the learned dynamic properties at inference time by applying explicit scaling factors $(\alpha, \beta, \gamma) \in \mathbb{R}^3$ to the initial velocity, potential gradient, and non-conservative force, respectively. The edited acceleration becomes:
\begin{equation}
\label{eq9}
\ddot{q} = -\beta \nabla_{q} U(q) + \gamma Q(q, \dot{q}),
\end{equation}
paired with the scaled initial state $(q^j_0, \alpha \dot{q}^j_0)$. Modulating these physical parameters allows us to seamlessly simulate counterfactual environments, such as scaling the initial velocity ($\alpha \neq 1$), weakening the effective field strength ($\beta < 1$), or adjusting the non-conservative dissipation/driving effects ($\gamma \neq 1$).

\begin{figure}
\centering
\includegraphics[width=0.9\linewidth]{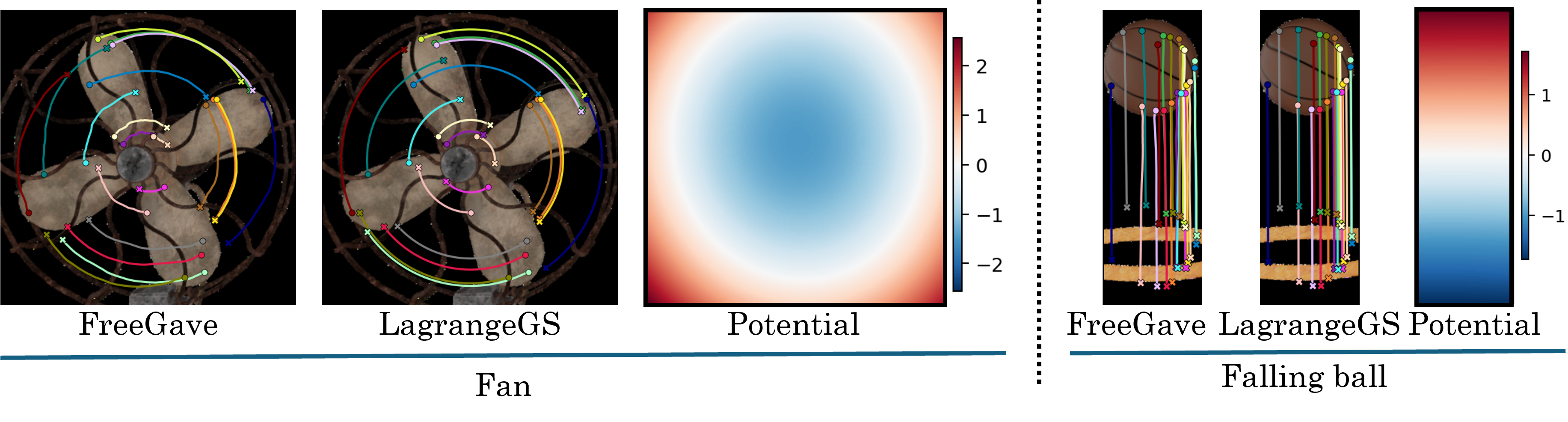}
\vspace{-4mm}
\caption{\label{fig4} Learned trajectories of FreeGave and LagrangeGS with the potential $U(q)$ for the "Fan" and "Falling Ball" scenes. While the rollouts of FreeGave exhibit erratic fluctuations, LagrangeGS ensures consistent trajectories satisfying Lagrangian mechanics. Without structural priors, $U(q)$ captures the underlying physics: a rotationally symmetric potential for the Fan, and a uniform gravitational gradient for the Falling Ball.}
\vspace{-5mm}
\end{figure}

\section{\label{sec4} Experiments}
\subsection{\label{sec4-1} Experimental setup}
\textbf{Datasets.} We evaluate LagrangeGS on two dynamic 3D scene benchmarks. The Dynamic Object dataset~\cite{li2023nvfi} contains six multi-view scenes featuring both globally rigid and locally deformable motions. The Dynamic Indoor Scene dataset~\cite{li2023nvfi} comprises four scenes in which multiple rigid objects move simultaneously against a static background. Both benchmarks expose held-out time windows to evaluate temporal extrapolation, in addition to held-out camera views for novel-view synthesis.\smallskip\\
\textbf{Baselines.} We compare LagrangeGS with two types of dynamic-scene methods. The first type models the scene as a deformation of an initial representation: T-NeRF~\cite{pumarola2021d}, D-NeRF~\cite{pumarola2021d}, NSFF~\cite{li2021neural}, and DefGS~\cite{yang2024deformable}. The second type additionally introduces physical intermediates: NVFi~\cite{li2023nvfi}, TRACE~\cite{li2025trace}, and FreeGave~\cite{li2025freegave}. All baselines are evaluated on the same train/test splits as the proposed method.\smallskip\\
\textbf{Pre-trained models.} LagrangeGS is a universal physical framework that extracts its initial kinematic trajectories from a pre-trained dynamic 3DGS model. To demonstrate its robustness against variations in initial kinematic quality, we apply LagrangeGS using two representative models: DefGS~\cite{yang2024deformable} (purely visual) and FreeGave~\cite{li2025freegave} (velocity-regularized). For each source, we train using the official implementation and default hyperparameters.

\begin{figure}
\centering
\includegraphics[width=0.95\linewidth]{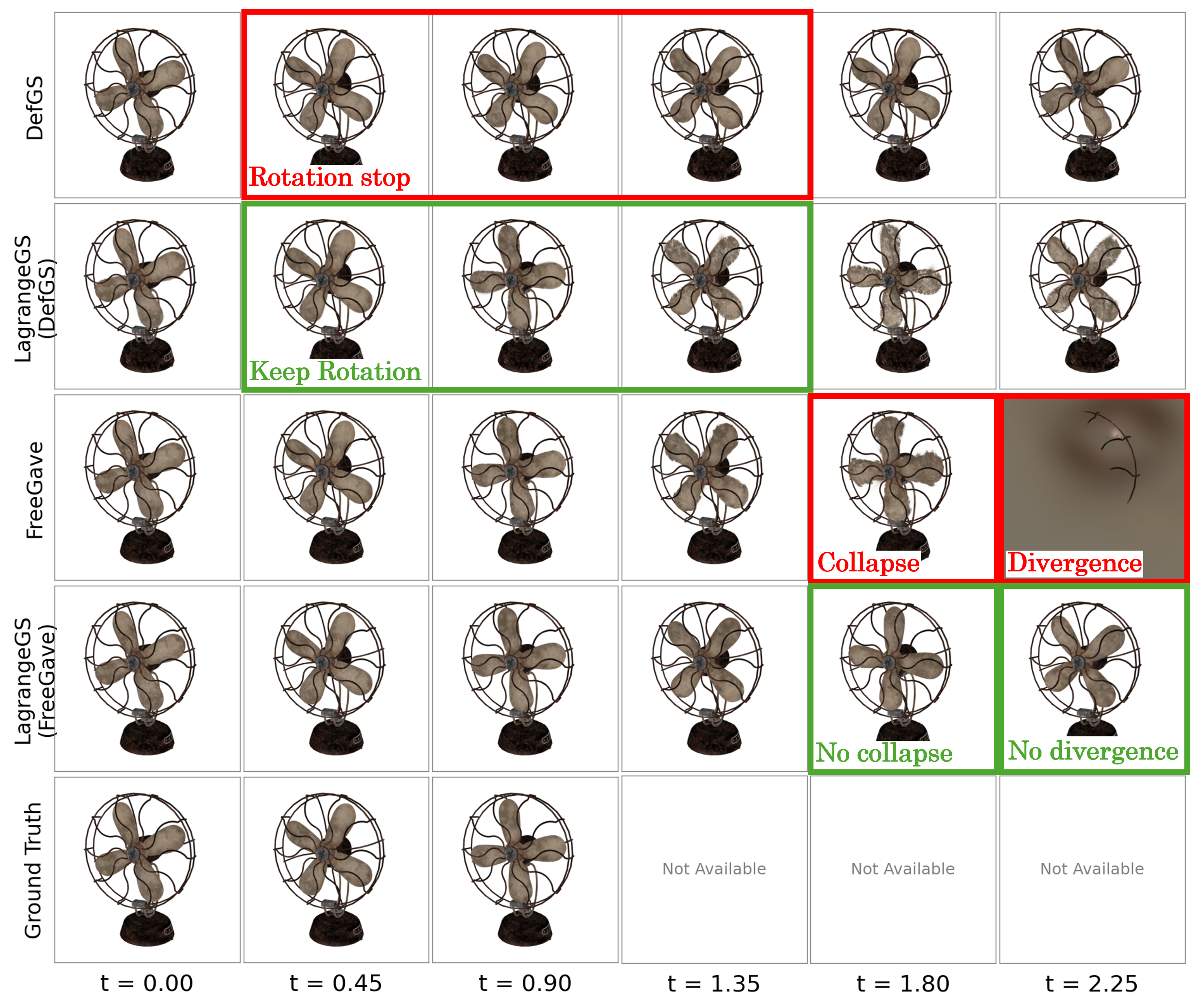}
\vspace{-3mm}
\caption{\label{fig5} Extrapolation on the Fan scene. Training spans $t = 0.0$–$0.75$. Upon entering extrapolation, DefGS instantly suffers from unnatural motion or complete stagnation, while FreeGave exhibits geometric drift at $t = 1.80$ and catastrophic volume divergence at $t = 2.25$. Distilling both backbones via LagrangeGS (Ours) mitigates these artifacts, sustaining physically consistent and smooth rotation across the entire extended horizon.}
\vspace{-5mm}
\end{figure}

\noindent\textbf{Evaluation metrics.} We measure photorealistic rendering quality using standard metrics: Peak Signal-to-Noise Ratio (PSNR), Structural Similarity Index Measure (SSIM), and Learned Perceptual Image Patch Similarity (LPIPS). We qualitatively assess the performance on long-term extrapolation, time reversal, and counterfactual physics-based editing.\smallskip\\
\textbf{Implementation details.} The potential energy $U$ and the non-conservative force $Q$ in the TD-LNN are predicted by four-layer MLPs with $128$ hidden units and Sigmoid-weighted Linear Unit (SiLU) activations. The last layer of $Q$ is zero-initialized to enforce a conservative dynamics baseline at the start of fine-tuning. We utilize a single NVIDIA DGX Spark for distillation and fine-tuning. Full hyperparameter settings and detailed network architectures are provided in the supplementary material. On a DGX Spark, the complete LagrangeGS pipeline requires 3.5 h for training and 90 s for testing, compared with 2.5 h and 60 s for FreeGave. Training memory increases from 4.7 GB to 9.6 GB due to differentiable multi-step rollout, while both methods require 3.3 GB during testing.

\subsection{\label{sec4-2-} Visual analysis of trajectories and potentials}
Before evaluating the extrapolation capabilities, we first visually analyze Gaussian particle trajectories and the learned potential fields on representative scenes. As shown in Fig.~\ref{fig4}, the FreeGave trajectories exhibit erratic fluctuations and fail to satisfy Lagrangian mechanics, while LagrangeGS yields physically consistent trajectories. Additionally, the learned potentials successfully capture the underlying scene physics without structural priors: recovering a rotationally symmetric harmonic potential for the Fan rotation, and a uniform gradient field strictly corresponding to gravitational acceleration for the Falling Ball.

\begin{figure}
\centering
\includegraphics[width=0.95\linewidth]{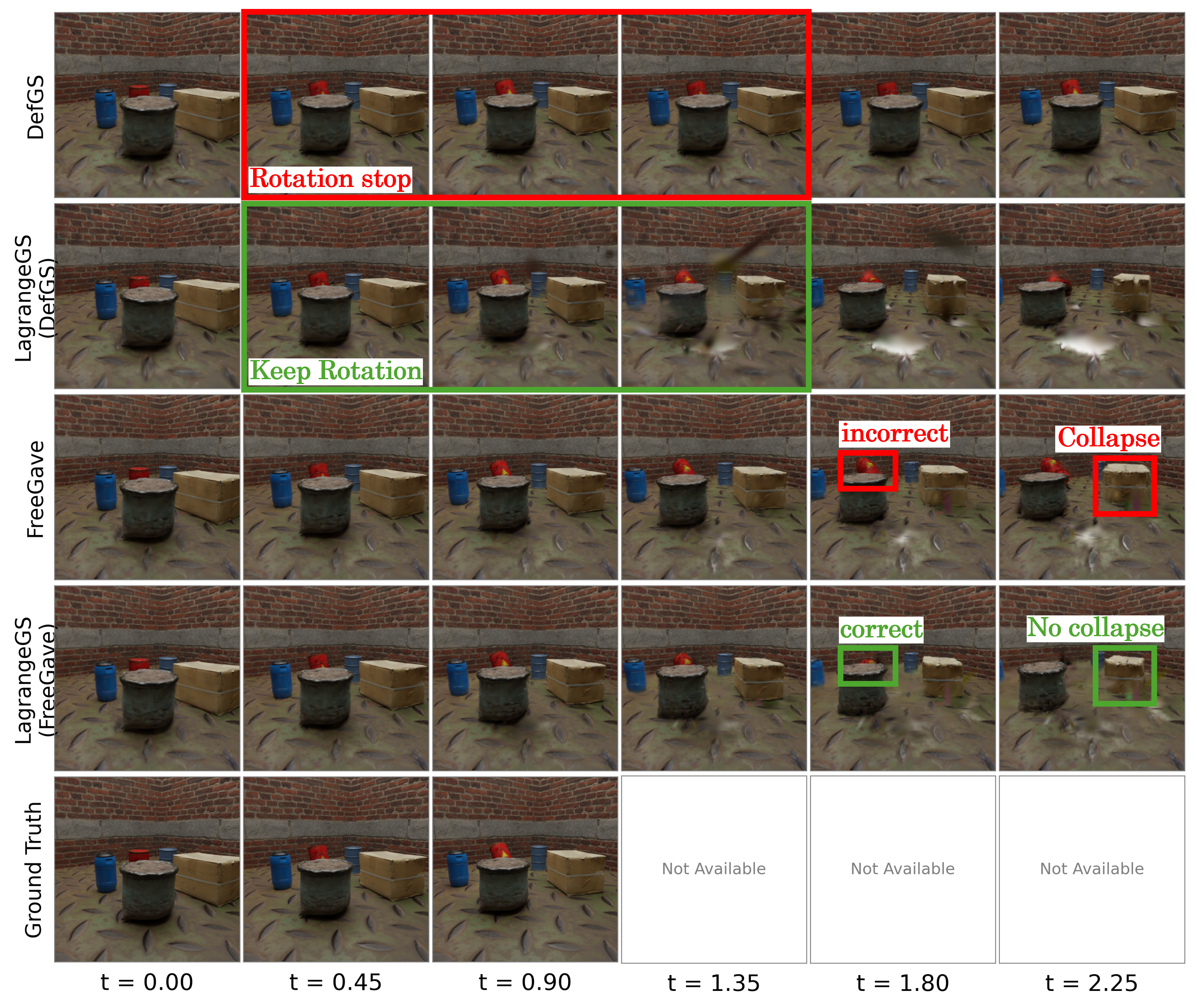}
\vspace{-3mm}
\caption{\label{fig7} Extrapolation on the Factory scene. Training spans $t = 0.0$–$0.75$. Upon extrapolation, DefGS immediately degenerates into an indistinct haze, while FreeGave suffers from geometric collapse for $t \ge 1.80$. LagrangeGS prevents this long-rollout collapse for both backbones, preserving the can geometry.}
\vspace{-5mm}
\end{figure}

\begin{table}
\centering
\caption{\label{tab1} Short-term extrapolation performance on dynamic scene benchmarks. LagrangeGS improves the extrapolation performance of DefGS. Conversely, when applied to FreeGave, photometric metrics exhibit a marginal drop. This is a plausible consequence of strict physical regularization: by satisfying Lagrangian mechanics across all particles, LagrangeGS deliberately prunes non-physical visual artifacts such as shadows.}
\resizebox{0.85\textwidth}{!}{
\begin{tabular}{r|ccc|ccc}
\toprule
&\multicolumn{3}{c|}{Dynamic Object Dataset}&\multicolumn{3}{c}{Dynamic Indoor Scene Dataset} \\
&PSNR$\uparrow$&SSIM$\uparrow$&LPIPS$\downarrow$&PSNR$\uparrow$&SSIM$\uparrow$&LPIPS$\downarrow$ \\ \midrule
T-NeRF~\cite{pumarola2021d}&13.818&0.739&0.324&22.242&0.700&0.363 \\
D-NeRF~\cite{pumarola2021d}&14.660&0.737&0.312&20.791&0.692&0.349 \\
NSFF~\cite{li2021neural}&-&-&-&24.163&0.795&0.289 \\
TiNeuVox~\cite{fang2022fast}&19.612&0.940&0.073&21.029&0.770&0.281 \\
NVFi~\cite{li2023nvfi}&27.594&0.972&0.036&29.745&0.876&0.204 \\
TRACE~\cite{li2025trace}&31.597&0.987&0.009&34.824&0.965&0.054 \\ 
\midrule
DefGS~\cite{yang2024deformable}&19.849  &0.949   &0.045   &21.380  &0.819   &0.188   \\
LagrangeGS (DefGS)   &24.900  &0.965   &0.037   &31.370  &0.930   &0.103   \\
diff                &(+5.051)&(+0.016)&(-0.008)&(+9.990)&(+0.111)&(-0.085)\\
\midrule
FreeGave~\cite{li2025freegave}&31.987  &0.990   &0.007   &35.019  &0.966   &0.051   \\
LagrangeGS (FreeGave)&28.940  &0.977   &0.020   &33.097  &0.955   &0.059   \\
diff                &(-3.047)&(-0.013)&(+0.013)&(-1.922)&(-0.011)&(+0.008)\\
\bottomrule
\end{tabular}
}
\end{table}

\subsection{\label{sec4-2} Extrapolation}
We evaluate the extrapolation capabilities of LagrangeGS and the baselines by rolling out the learned dynamics beyond the training window ($t > t_{\rm{max}}$). The results are visualized in Fig.~\ref{fig5} and Fig.~\ref{fig7}. Note that the training window spans $t = 0.0$–$0.75$; hence, we consider $t = 0.0$–$0.75$ as interpolation, $t = 0.75$–$1.0$ as short-term extrapolation, and $t > 1.0$ as long-term extrapolation. DefGS suffers from unnatural motion immediately upon extrapolation, while FreeGave exhibits geometric collapse and volume divergence at later time steps. In contrast, LagrangeGS (DefGS) preserves physically consistent and smooth motion across the entire extended horizon. LagrangeGS (FreeGave) also maintains stable motion and prevents the artifacts observed in the original FreeGave.

Quantitatively, as shown in Table~1, LagrangeGS improves the photometric metrics when applied to DefGS, whereas a moderate degradation is observed when initialized from FreeGave. To isolate whether this degradation originates from appearance or geometry, we additionally compute PSNR on binarized foreground silhouettes. On Dynamic Object, the FreeGave--LagrangeGS gap is nearly unchanged between RGB and silhouette PSNR ($-3.2$ vs.\ $-3.2$ dB), indicating that the degradation is primarily geometric. In contrast, on Dynamic Indoor Scene, the gap decreases substantially from $-2.5$ dB in RGB to $-0.6$ dB in silhouette PSNR, indicating that the degradation is mainly caused by appearance effects such as moving shadows.

To evaluate long-term extrapolation under constrained conditions, we artificially truncate the training window to $t < 0.5$ and $t < 0.3$ for FreeGave and LagrangeGS (FreeGave), and evaluate their performance during the original extrapolation window ($0.75 < t \le 1.0$). As shown in Table~\ref{tab2}, the performance gap between the two models narrows as the training window is shortened. Especially under extreme truncation ($t < 0.3$) in the Dynamic Object dataset, LagrangeGS outperforms FreeGave, demonstrating its resilience to temporal overfitting and its structural stability in data-sparse regimes.

To quantitatively evaluate geometric stability during long-term extrapolation, we measure the foreground pixel-area extent relative to that at $t=0$, since unstable rollouts typically manifest as object expansion or divergence. On the Dynamic Object dataset, LagrangeGS remains within $1.1\times$ its initial foreground extent throughout the rollout, whereas FreeGave expands up to $1.9\times$. This result quantitatively confirms that LagrangeGS suppresses the geometric divergence observed in long-term rollout.

\subsection{\label{sec4-3} Time reversal}
\begin{figure}
\centering
\includegraphics[width=1.0\linewidth]{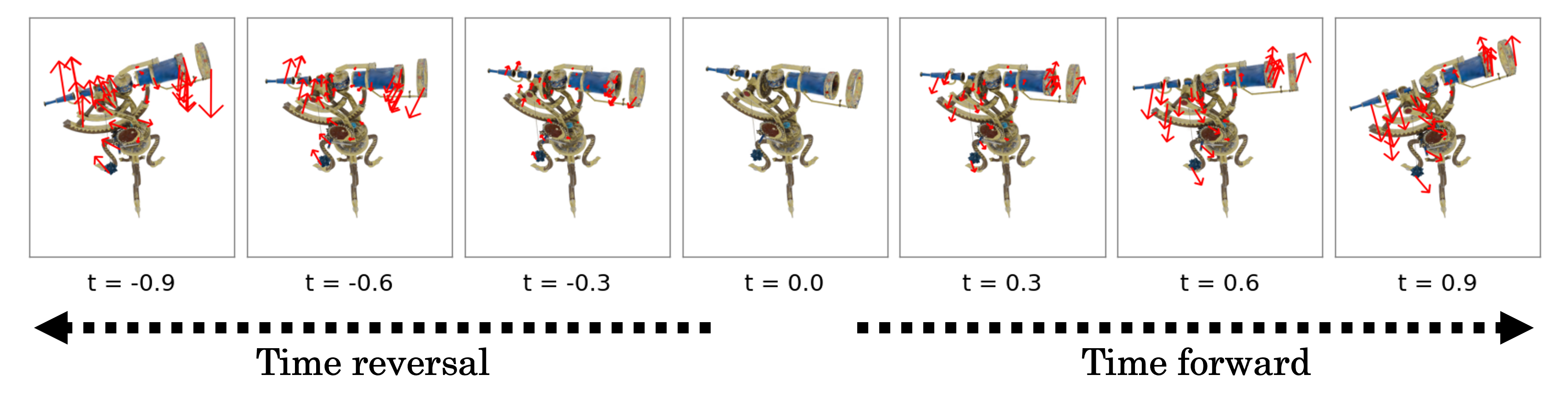}
\vspace{-3mm}
\caption{\label{fig8} Time-reversal rollout on the "Telescope" scene via LagrangeGS. Past states ($t < 0$) are recovered exactly by integrating the same trained TD-LNN backward with a negative time step $-\Delta t$, ensuring visual and structural consistency with forward rollouts ($t > 0$). Red arrows denote per-particle velocity vectors; their exact directional inversion across $t=0$ confirms that LagrangeGS captures genuine continuous mechanics rather than a learned animation shortcut.}
\end{figure}

We validate the time-reversal capability of LagrangeGS by rolling out the learned TD-LNN backward in time with a negative time step $-\Delta t$. As verified by the per-particle velocity vectors shown in Fig.~\ref{fig8}, the velocity directions invert exactly across the temporal boundary ($t=0$). Consequently, the recovered past states at $t = -0.3, -0.6, -0.9$ exhibit structural and visual consistency with the corresponding forward rollout trajectories at $t = +0.3, +0.6, +0.9$. To quantify this consistency, we perform a forward--backward cycle test by integrating each trajectory from $t=0$ to $t_{\max}$ and subsequently back to $t=0$. We measure the final position error $\|\mathbf{q}_0-\hat{\mathbf{q}}_0\|$ normalized by the total traveled distance. The mean cycle error is only 1.3\% on Dynamic Object and 0.19\% on Dynamic Indoor Scene, demonstrating strong forward--backward consistency despite the absence of an exact time-reversal guarantee.

\begin{table}
\centering
\caption{\label{tab2} Extrapolation under truncated observation windows ($t<0.5$, $t<0.3$). By artificially shortening the training horizon (originally $t_{max} \sim 0.75$), the performance gap between FreeGave and LagrangeGS narrows as the training window is shortened. Especially for the extreme truncation ($t < 0.3$) in the Dynamic Object dataset, LagrangeGS outperforms FreeGave, demonstrating its resilience in long-term extrapolation under data-sparse regimes.}
\resizebox{0.85\textwidth}{!}{
\begin{tabular}{c|c|ccc|ccc}
\toprule
&&\multicolumn{3}{c|}{Dynamic Object Dataset}&\multicolumn{3}{c}{Dynamic Indoor Scene Dataset} \\
&&PSNR$\uparrow$&SSIM$\uparrow$&LPIPS$\downarrow$&PSNR$\uparrow$&SSIM$\uparrow$&LPIPS$\downarrow$ \\ \midrule
       &FreeGave  &31.987  &0.990   &0.007   &35.019  &0.966   &0.051   \\
Full   &LagrangeGS&28.940  &0.977   &0.020   &33.097  &0.955   &0.059   \\
       &diff      &(-3.047)&(-0.013)&(+0.013)&(-1.922)&(-0.011)&(+0.008)\\
\midrule
       &FreeGave  &24.874  &0.970   &0.027   &25.757  &0.823   &0.214   \\
$t<0.5$&LagrangeGS&24.601  &0.962   &0.032   &24.713  &0.821   &0.218   \\
       &diff      &(-0.273)&(-0.008)&(+0.005)&(-1.044)&(-0.002)&(+0.004)\\
\midrule
       &FreeGave  &21.341  &0.960   &0.038   &23.677  &0.789   &0.246   \\
$t<0.3$&LagrangeGS&22.272  &0.949   &0.042   &22.975  &0.786   &0.246   \\
       &diff      &(+0.913)&(-0.011)&(+0.004)&(-0.702)&(-0.003)&(+0.000)\\
\bottomrule
\end{tabular}
}
\end{table}

\subsection{\label{sec4-4} Counterfactual physics-based editing}
We demonstrate the counterfactual physics-based editing capability of LagrangeGS by directly rescaling the learned potential field $U$ at inference time. As shown in Fig.~8, decreasing the scaling factor $\beta$ weakens the effective field and slows the ball's descent, whereas increasing $\beta$ accelerates the motion.

We further quantify whether this manipulation follows the intended physical change. For the Falling Ball scene, scaling the learned potential by $\beta$ changes the falling time according to the expected relation $t\propto\beta^{-1/2}$. Fitting the simulated falling times yields $t\propto\beta^{-0.489}$ with $R=0.999$, closely matching the theoretical exponent of $-0.5$. This demonstrates that the learned potential supports quantitatively calibrated physics-based editing rather than merely producing visually plausible trajectory changes.

LagrangeGS also supports scaling the initial velocity and the non-conservative force, enabling counterfactual simulations such as varying the initial impulse or dissipation/driving effects.

\begin{table}
\centering
\caption{\label{tab3} Ablation study. The Stage 2 (TD-LNN) and Stage 3 (Gaussian parameter) fine-tuning contribute to visual scores. While LRA is essential for preventing geometric collapse, its absence marginally inflates PSNR in the Indoor dataset; this merely reflects unconstrained particles overfitting to non-rigid visual phenomena like moving shadows. Furthermore, because the Indoor scenes exhibit strictly conservative motions, the non-conservative force $Q$ is redundant, meaning its removal avoids over-parameterization and slightly benefits PSNR.}
\resizebox{0.85\textwidth}{!}{
\begin{tabular}{c|ccc|ccc}
\toprule
&\multicolumn{3}{c|}{Dynamic Object Dataset}&\multicolumn{3}{c}{Dynamic Indoor Scene Dataset} \\
&PSNR$\uparrow$&SSIM$\uparrow$&LPIPS$\downarrow$&PSNR$\uparrow$&SSIM$\uparrow$&LPIPS$\downarrow$ \\ \midrule
Full        &28.940  &0.977   &0.020   &33.097  &0.955   &0.059   \\
w/o stage3  &27.608  &0.975   &0.018   &27.804  &0.934   &0.079   \\
w/o stage2,3&27.471  &0.974   &0.021   &29.011  &0.938   &0.077   \\
w/o LRA     &26.964  &0.974   &0.022   &34.261  &0.957   &0.057   \\
w/o Force Q &24.971  &0.965   &0.032   &33.303  &0.956   &0.054   \\
\bottomrule
\end{tabular}
}
\end{table}

\subsection{\label{sec4-5} Ablation studies}
We perform an ablation study for key components of LagrangeGS. Table~\ref{tab3} summarizes the results. TD-LNN fine-tuning (Stage 2) and Gaussian parameter refinement (Stage 3) consistently improve rendering fidelity. LRA is essential for preventing geometric collapse during long-term extrapolation and improving visual scores in the Dynamic Object dataset. However, its absence marginally inflates the scores in the Indoor dataset. The lack of LRA allows the particles to overfit to non-rigid visual phenomena, such as moving shadows, within the expressivity limits of $U, Q$, and $\dot{q}_0$. The non-conservative force $Q$ significantly benefits the Dynamic Object dataset, which contains non-conservative motions; however, it is redundant for the strictly conservative motions in the Indoor dataset, where its removal avoids over-parameterization and slightly benefits photometric scores.

\subsection{\label{sec4-6} Limitations}
While LagrangeGS successfully constrains dynamic 3DGS within Lagrangian mechanics, it exhibits two key limitations. First, the current TD-LNN requires distillation from a pre-trained dynamic 3DGS model, since the unstable nature of the initial random trajectories makes it difficult to directly train from scratch. Future work could explore more robust training strategies or curriculum learning to enable direct end-to-end training. Second, the current formulation assumes decoupled particle dynamics, which may not capture complex interactions such as collisions or contact forces. Extending the framework to incorporate more sophisticated kinetic priors or interaction models could further enhance its applicability to a wider range of dynamic scenes.

\begin{figure}
\centering
\includegraphics[width=0.95\linewidth]{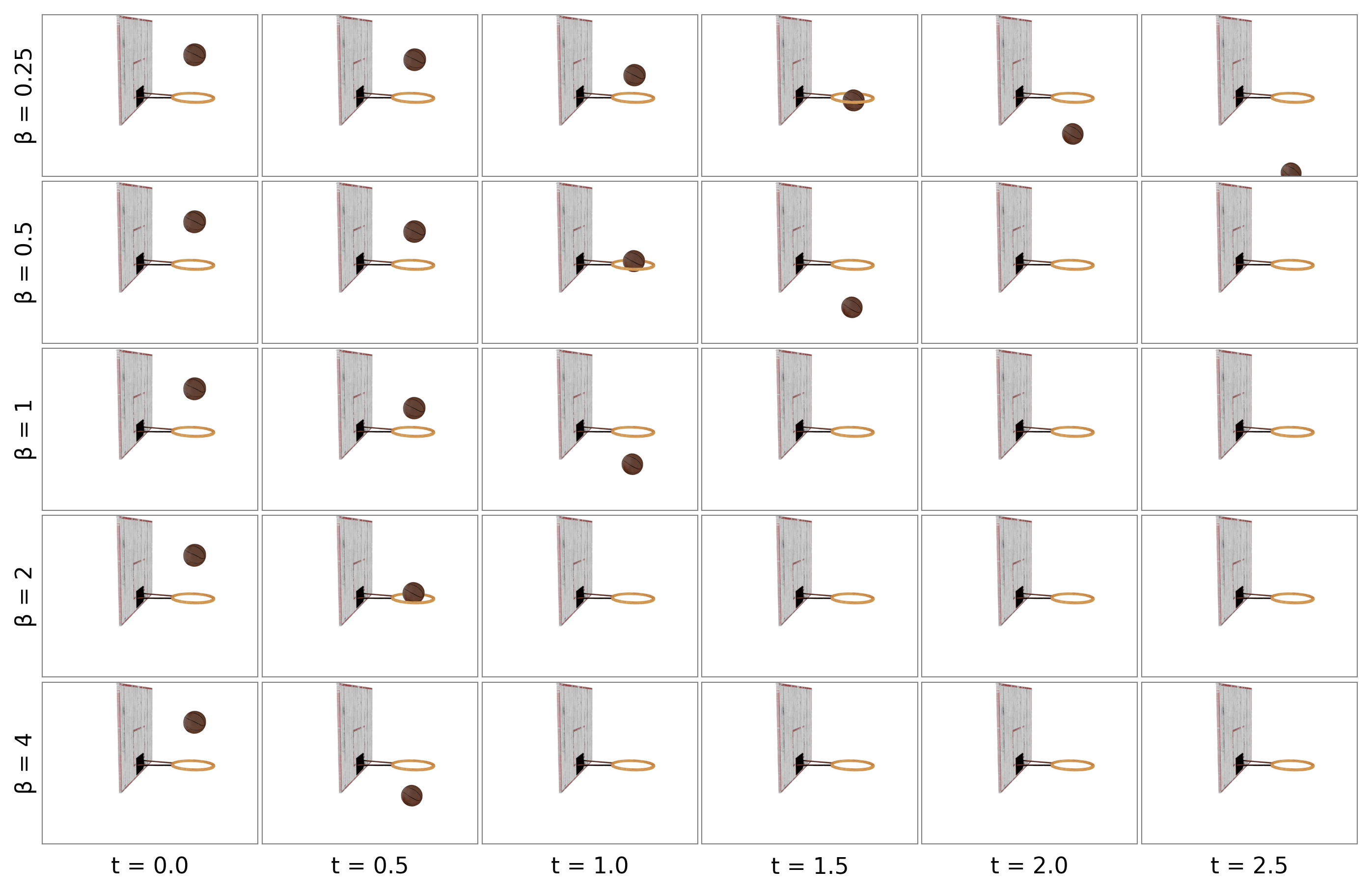}
\vspace{-3mm}
\caption{\label{fig9}  Inference-time counterfactual physics-based editing on the "Falling Ball" scene. The dynamics are systematically manipulated without retraining by rescaling the learned potential field ($U' = \beta \cdot U$). The baseline ($\beta = 1.0$) reproduces the true observed trajectory. Decreasing $\beta$ simulates a weakened gravitational environment, slowing the descent, while increasing $\beta$ strengthens the effective gravity field. This predictable and continuous modulation exposes the learned $U(q)$ as an explicitly editable physical property rather than an uninterpretable visual latent feature.}
\vspace{-5mm}
\end{figure}

\section{\label{sec5} Conclusion}
In this paper, we introduced LagrangeGS, a novel framework that integrates Lagrangian mechanics into dynamic 3DGS. LagrangeGS features two key components: an TD-LNN, which approximates the velocity-Hessian as an identity matrix for computational tractability and employs a time-independent system for consistent time reversal; and LRA, which binds the decoupled particles into locally rigid structures to prevent geometric collapse. Our experiments on two dynamic scene benchmarks demonstrate that LagrangeGS improves long-term extrapolation and enables consistent time reversal and counterfactual physics-based editing. This work establishes an explicit formulation of dynamic scenes as Lagrangian systems, paving the way for physically grounded dynamic modeling and manipulation in 3DGS.

\bibliography{egbib}

\end{document}